\documentclass{article}

\usepackage{microtype}
\usepackage{graphicx}
\usepackage{subcaption}
\usepackage{booktabs}
\usepackage{hyperref}

\usepackage[accepted]{icml2026}

\makeatletter
\renewcommand{\Notice@String}{%
  \textit{Accepted at the ICML 2026 Workshop on
  Resource-Adaptive Foundation Model Inference (AdaptFM), Seoul, South Korea.}%
}
\makeatother

\hypersetup{
  pdfsubject={Accepted at the ICML 2026 Workshop on Resource-Adaptive Foundation Model Inference (AdaptFM)}
}

\usepackage{amsmath}
\usepackage{amssymb}
\usepackage{mathtools}
\usepackage{amsthm}
\usepackage[capitalize,noabbrev]{cleveref}
\usepackage{tabularx}
\usepackage{colortbl}
\usepackage{makecell}
\usepackage{enumitem}
\usepackage{xurl}
\usepackage{xspace}
 
\renewcommand{\paragraph}[1]{\vspace{0.1em}\noindent\textbf{#1}}

\theoremstyle{plain}

\theoremstyle{definition}

\theoremstyle{remark}

\newcommand{\newrecipe}{Mosaic NVFP4\xspace}  
\newcommand{\nvrecipe}{NVIDIA '25 NVFP4\xspace}  

\icmltitlerunning{What Matters for NVFP4 Training?}

\begin{document}

\twocolumn[
  \icmltitle{What Matters for NVFP4 Training?%
  \texorpdfstring{\\}{ }%
  A Comparative Study of Low-Precision Pretraining Recipes}

  \begin{icmlauthorlist}
    \icmlauthor{Anjulie Agrusa}{nv}
    \icmlauthor{Andrei Panferov}{ista}
    \icmlauthor{Elizabeth Wei}{nv}
    \icmlauthor{Keith Wyss}{nv}
    \icmlauthor{Paul Gibbons}{nv}
    \icmlauthor{Erik Schultheis}{ista}
    \icmlauthor{Jinhang Choi}{nv}
    \icmlauthor{Frank Sun}{nv}
    \icmlauthor{Tijmen Blankevoort}{nv}
    \icmlauthor{Dan Alistarh}{ista,redhat}
  \end{icmlauthorlist}

  \icmlaffiliation{nv}{NVIDIA}
  \icmlaffiliation{ista}{Institute of Science and Technology Austria}
  \icmlaffiliation{redhat}{Red Hat AI}
  \icmlcorrespondingauthor{Dan Alistarh}{dan.alistarh@ist.ac.at}
  \icmlkeywords{low-precision training, NVFP4, quantization, large language models}

  \vskip 0.3in
]

\printAffiliationsAndNotice{}

\begin{abstract}
Training large language models directly in 4-bit floating-point (FP) formats promises substantial improvements in throughput and energy efficiency. While some recipe design choices have been validated at scale, many promising approaches remain untested beyond small models and short token horizons, leaving open the question of which trends will hold. We present a systematic comparison of recent NVFP4 recipes at medium model scale, evaluating dense 8B and 30B-A3B MoE models trained for up to 1T tokens, with a focus on which ingredients are necessary for accuracy recovery. We propose a final recipe grounded in the principles behind established stable NVFP4 training at scale, incorporating state-of-the-art techniques such as unbiased gradient estimation with lower quantization error than stochastic rounding. To the best of our knowledge, this is the strongest FP4 training result demonstrated at this scale to date in loss gap to BF16. Through ablation studies, we find that: (i) each technique in the optimized recipe measurably improves loss trajectory, (ii) selective high-precision layers are necessary for recovering accuracy at scale, (iii) not all tensors in the backward pass benefit equally from de-biasing, leaving room to apply complementary error-reduction techniques to the remaining tensors.  
\end{abstract}

\vspace{-1em}
\section{Introduction}

The computational cost of training state-of-the-art foundation models has been increasing rapidly~\citep{amodei2018ai,sevilla2022compute}. Since pretraining modern Transformer-based models is dominated by dense matrix multiplications, reducing the precision of these operations is one of the most direct levers for improving training efficiency. This has motivated a progression from FP16/BF16 to FP8 training \citep{micikevicius2022fp8}, and, more recently, to hardware-supported 4-bit block scale formats \citep{rouhani2023microscalingdataformatsdeep} such as MXFP4 and NVFP4 on NVIDIA Blackwell GPUs \citep{nvidia2025blackwell, alvarez2025introducingnvfp4, nvidia2025nvfp4}.

Yet, quantized training introduces noise, and in the 4-bit regime this noise could affect convergence, even when operations may appear accurate. A recent line of work has therefore focused on recipes that recover most of the accuracy of FP8 or BF16 training while preserving the speedups of 4-bit tensor-core execution. Examples include the \nvrecipe recipe \citep{nvidia2025pretraining}, backward-pass FP4 training \citep{tseng2025mxfp4}, fully quantized FP4 training \citep{chmiel2025fp4all}, TetraJet \citep{chen2025oscillationreducedmxfp4trainingvision}, TetraJet-v2 \citep{chen2025tetrajet}, Four Over Six \citep{cook2025fouroversix}, Quartet \citep{castro2025quartet}, and Quartet II \citep{panferov2026quartetii}. Despite this rapid progress, it is fair to say that there is still no definitive solution: low-precision training can still induce small but persistent accuracy degradation, and existing methods disagree about which techniques are essential.

At the same time, some principles are starting to emerge. On the forward pass, multiple methods implicitly or explicitly optimize representation quality, for instance by substituting the \nvrecipe{} recipe's two-dimensional weight quantization \citep{nvidia2025pretraining} with one-dimensional quantization for finer-grained scaling or by reducing quantization error for weights and activations through improved scale selection \citep{chen2025oscillationreducedmxfp4trainingvision, castro2025quartet, chen2025tetrajet, cook2025fouroversix, panferov2026quartetii}. On the backward pass, a line of evidence suggests that bias in the gradient estimator can accumulate over long training horizons, making unbiased or carefully controlled quantization important for stable convergence \citep{chmiel2024accurateneuraltraining4bit, tseng2025mxfp4, chen2025oscillationreducedmxfp4trainingvision, chmiel2025fp4all, nvidia2025pretraining, chen2025tetrajet, panferov2026quartetii}. The difficulty is that these principles are not always clearly compatible. For instance, the \nvrecipe recipe uses two-dimensional quantization of weights to maintain consistency between forward and backward quantized weight representations \citep{nvidia2025pretraining}. By contrast, Quartet II introduces a technique to address bias in the backward pass that relies on random rotations applied along the inner dimension of both backward GEMMs; those rotations, in turn, are most naturally paired with one-dimensional quantization \citep{panferov2026quartetii, tseng2025mxfp4, chen2025tetrajet}. To approximate consistency with the forward-pass weight representation, one-dimensional schemes often dequantize the forward-quantized weights and re-quantize them along the backward contraction dimension rather than directly quantizing the high-precision weights during the backward pass \citep{chen2025oscillationreducedmxfp4trainingvision, castro2025quartet, chen2025tetrajet, panferov2026quartetii}.

This creates a practical question: which ingredients are non-negotiable, and which ones are merely useful at a given scale? The key problem we aim to address here is that the ``right'' answer is hard to infer from short ablations. Many accuracy effects in low-precision training only become visible late in training, or only after increasing model scale and token budget. As a result, as we show, recipes can appear close early on while diverging later, in ways that are expensive to diagnose. 

In this paper, we aim to isolate a set of  principles that matter for NVFP4 training by systematically comparing variants of existing recipes at medium scale. We evaluate dense 8B and 30B-A3B mixture-of-experts (MoE) models from the Nemotron  model family, trained up to 1T tokens. We organize our comparison around the major design axes that separate current methods: whether to use selective high-precision layers, how to select scales in the forward and backward pass, whether to quantize weights in one or two dimensions, strategies to mitigate bias in the backward pass, and how these choices interact.

\paragraph{Contributions.}
Our main finding is that the current recipes capture complementary pieces of the problem, and we propose a ``Mosaic'' approach that blends the best qualities of each. In more detail:
\begin{enumerate}
    \item We perform the first detailed and unified ablation of NVFP4 training techniques, examining how each technique reduces the model-quality gap relative to BF16 training.  
    \item  
    We show that keeping selected layers in higher precision is necessary for    
  full accuracy recovery in NVFP4 training, even with an optimized recipe. Omitting this can cause training instabilities at larger model sizes and medium-scale token horizons.    
    \item 
     We find that selectively applying the MS-EDEN \citep{panferov2026quartetii} unbiased estimator to gradients and Four Over Six \citep{cook2025fouroversix} scale selection to the remaining backward tensors yields the lowest observed loss gap among the tested backward-pass configurations when all layers are quantized.
    \item 
    Based on these observations, we develop a combined NVFP4 training recipe, which we call ``Mosaic NVFP4,'' with selective high-precision layers, Random Hadamard Transforms on both backward GEMMs, unbiased gradients via MS-EDEN, proposed by Quartet II, and Four Over Six scale selection on remaining forward and backward tensors. To the best of our knowledge, this is the strongest FP4 training result demonstrated at this scale in loss gap to BF16. This scheme remains compatible with efficient kernels. 
\end{enumerate}

\section{Related Work}

\paragraph{Low-Precision training.}
Low-precision training has a long history, including early work on reduced-precision and integer training for smaller neural networks \citep{courbariaux2015binaryconnect,banner2018scalable,yang2019training}. For Transformer models, 8-bit training has become increasingly practical through INT8 methods such as SwitchBack and JetFire, FP8 training systems, and 8-bit LLM fine-tuning methods such as HALO \citep{wortsman2023stablelowprecisiontraininglargescale, xi2024jetfireefficientaccuratetransformer, micikevicius2022fp8, deepseekai2025deepseekv3technicalreport, mishra2025recipespretrainingllmsmxfp8, ashkboos2025halohadamardassistedlowerprecisionoptimization}. These methods highlighted several recurring themes: the need for fine-grained scaling, special treatment of outlier values, and careful handling of the backward pass.

\paragraph{Training in FP4 formats.}
The introduction of hardware-supported FP4 microscaling formats has renewed interest in end-to-end 4-bit training. Early FP4 and INT4 work demonstrated that 4-bit arithmetic could be incorporated into training \citep{xi2023trainingtransformers4bitintegers, chmiel2024accurateneuraltraining4bit, wang2025optimizinglargelanguagemodel, zhou2025efficientpretrainingexploringfp4}. More recent work targets hardware-supported MXFP4 and NVFP4 training. \citet{tseng2025mxfp4} studied backward-pass quantization to FP4 using Hadamard rotations and stochastic rounding. The TetraJet series combined one-dimensional weight quantization with re-quantization in the backward pass and stochastic rounding, and developed methods for oscillation suppression; TetraJet-v2 extended the approach to NVFP4 LLM training, adding double-block quantization, rotations to both backward GEMMs, and an outlier control mechanism \citep{chen2025oscillationreducedmxfp4trainingvision, chen2025tetrajet}. \citet{chmiel2025fp4all} studied fully quantized FP4 training with NVFP4-style scaling and identified stochastic rounding as important for the backward and update passes. \citet{nvidia2025pretraining} introduced a large-scale NVFP4 recipe based on Random Hadamard transforms, two-dimensional weight quantization, stochastic rounding, and selective high-precision layers. Four Over Six \citep{cook2025fouroversix} improved NVFP4 quantization accuracy by adaptively choosing between two scale factors.

\begin{figure*}[h!]
\centering
  \includegraphics[width=\textwidth]{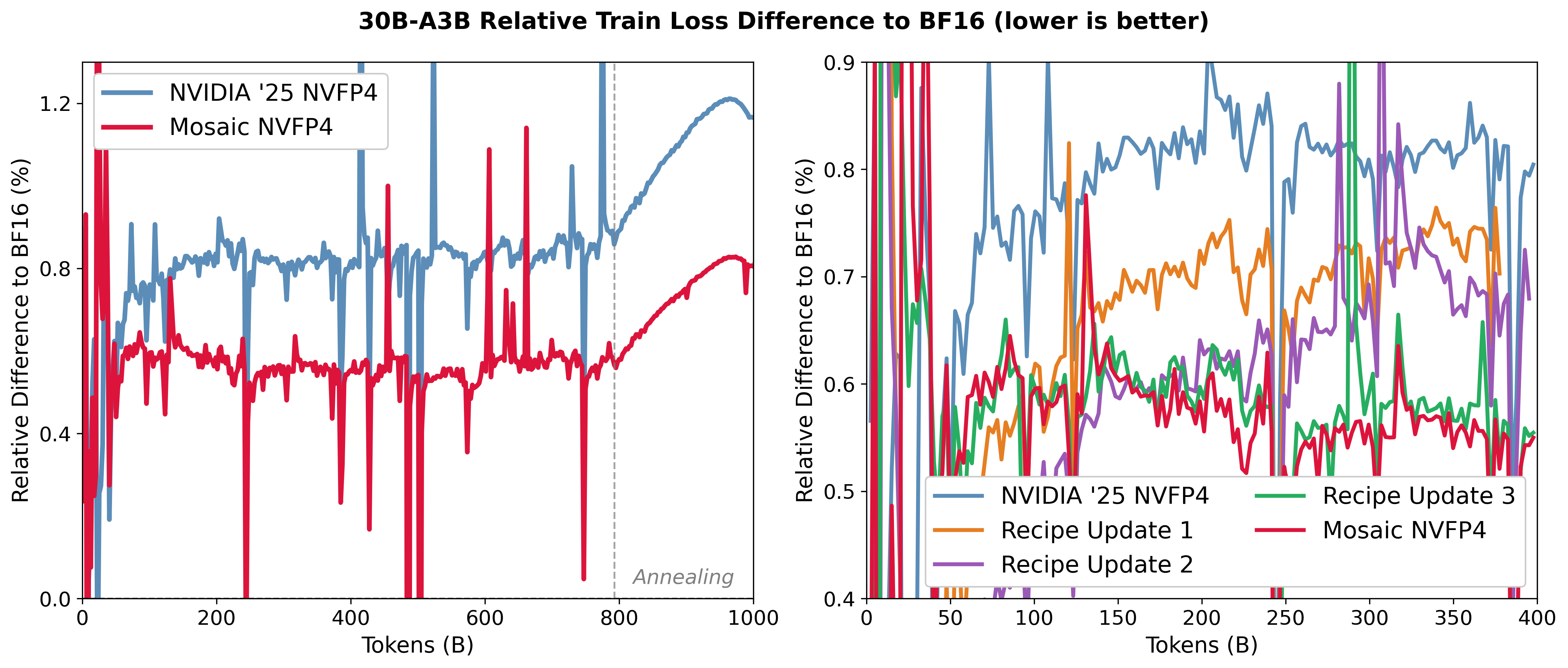}
  \caption{Left: Relative train loss difference between each NVFP4 recipe and BF16 (lower is better), trained with \nvrecipe and \newrecipe recipes to 1T tokens on Nemotron 3 Nano. Right: Cumulative recipe updates, each adding one \newrecipe component, trained to 400B tokens on Nemotron 3 Nano. All recipes keep the last 15\% of layers in BF16. Recipe updates are described in Table~\ref{recipe_table}.}
  \label{fig:recipe_stacking}
\end{figure*}

\paragraph{Quartet and unbiased gradient estimation.}
Quartet \citep{castro2025quartet} studied MXFP4 training through the lens of ``low-precision scaling laws,'' linking forward compression error to parameter efficiency and backward-pass bias to data efficiency. Historically, methods addressing bias in quantized gradient estimates have generally relied on elementwise stochastic rounding \citep{chmiel2024accurateneuraltraining4bit, tseng2025mxfp4, chen2025oscillationreducedmxfp4trainingvision, chmiel2025fp4all, nvidia2025pretraining, chen2025tetrajet}. Quartet II \citep{panferov2026quartetii} extended the Quartet line of work to NVFP4 and introduced MS-EDEN, which instead combines random rotations with a per-block corrective rescaling to produce lower-error unbiased estimates than standard FP4 stochastic rounding. Our work is closest in spirit to these studies, but asks which ingredients of existing FP4 training recipes matter most, and how they should be combined, when evaluated under a common implementation and training budget.

\section{Experimental Setup}
 All NVFP4 variants are implemented in a common training stack and trained on the same GPU architecture.

\paragraph{Models and data.}
We evaluate both dense and sparse hybrid
Mamba-Transformer architectures. The dense setting uses the 8B variant of the NVIDIA Nemotron-H Family of Hybrid Mamba-Transformer Models \citep{nvidia2025nemotronhfamilyaccurateefficient}. This model has 24 Mamba-2 layers, 4 Attention layers, and 24 FFN layers.  We train on the same data distribution as \citet{nvidia2025nemotronhfamilyaccurateefficient}, following their phased data-blending approach \citep{feng2024maximizedataspotentialenhancing} with diverse data for the first 600B tokens and high-quality data for the remainder of the 1T token horizon. 
 We additionally validated NVFP4 pretraining recipes on the Nemotron 3 Nano hybrid Mamba-Transformer MoE architecture (30B-A3B), using the first 1T tokens from the Phase 1 data mixture of the 25T-token pretraining curriculum described by \citet{nvidia2025nemotron3nanoopen, nvidia2025nvidianemotron3efficient}. Like Nemotron-H, this model has a high ratio of Mamba-2 to attention layers. Each MoE layer contains 128 routed experts, with top-6 routing and 2 shared experts. We report training and validation loss alongside downstream evaluations spanning knowledge, reasoning, math, and coding benchmarks.

\paragraph{Quantization method terminology.}
Quantization is applied to the network's linear layers and grouped linear layers. To describe quantization recipes, we refer to three general matrix-matrix multiplication (GEMM) operations and the quantized operands to these products. Assuming GEMM contracts along the last dimension of the two inputs, we formalize them as follows:
\begin{enumerate}
    \item \textbf{Y=GEMM(X, W)} --- the product between the input, X, and the weight, W.
    \item \textbf{DGRAD=GEMM(G, W.T)} --- the product between the gradient, G with respect to Y, and the transposed weight, W.T.
    \item \textbf{WGRAD=GEMM(G.T, X.T)} --- the product between the transposed gradient, G.T, and the transposed input, X.T.
\end{enumerate}

When both operands of a GEMM are quantized to NVFP4 along their last dimension, the multiplication can be performed using specialized Tensor Cores on Blackwell GPUs. As all operands across the three GEMMs are unique, we will use the notation above to discriminate between each tensor in a computation scheme.

\paragraph{Compared techniques.}
We instantiate the main design choices from prior recipes in a common framework:

\begin{itemize}
    \item \textbf{NVIDIA~'25 NVFP4:} two-dimensional weight quantization, stochastic rounding on gradients, static Random Hadamard Transforms (RHTs) on WGRAD, and selective high-precision layers \citep{nvidia2025pretraining}.

    \item \textbf{Quartet-II-style backward pass:} constructed through three cumulative updates \cite{panferov2026quartetii}:
    \begin{itemize}
        \item One-dimensional weight quantization with backward re-quantization \citep{chen2025oscillationreducedmxfp4trainingvision, castro2025quartet, chen2025tetrajet};
        \item Random Hadamard Transforms on both backward GEMMs \citep{tseng2025mxfp4, castro2025quartet, chen2025tetrajet}; and
        \item MS-EDEN applied to backward tensors \citep{panferov2026quartetii}.
    \end{itemize}
    
    Since the first two components have been studied as design choices in existing low-precision training recipes, we ablate them separately to measure their individual contributions to model quality. MS-EDEN adapts EDEN’s randomized-rotation and corrective-rescaling framework to microscaling formats \citep{vargaftik2022edencommunicationefficientrobustdistributed, panferov2026quartetii}. To instantiate these randomized rotations, we replace the baseline recipe's static RHT sign vectors with resampled vectors throughout training.

    \item \textbf{Four Over Six:} adaptive block-scale selection applied selectively \citep{cook2025fouroversix}.
\end{itemize}

\newcolumntype{Y}[1]{>{\hsize=#1\hsize\raggedright\arraybackslash}X}

\begin{table*}[h!]
  \centering
  \caption{Recipe improvements over the prior state of the art at this model and token scale.
    All recipes keep the last 15\% of layers in BF16 unless otherwise noted.}
  \label{recipe_table}
  \begin{tabularx}{\textwidth}{p{1.7cm} Y{0.7} Y{0.5} Y{1.8} Y{0.6} Y{0.8} Y{0.7}}
    \toprule
    \rowcolor{gray!15}
    \textbf{Recipe} &
    \textbf{Weight Quant} &
    \textbf{Weight Re-Quant} &
    \textbf{Tiled Random Hadamard Transforms} &
    \textbf{Random Sign Vectors} &
    \textbf{Technique to Address Bias in Gradients} &
    \textbf{Four Over Six Tensors} \\
    \midrule
    \textbf{\nvrecipe} &
    2D &
    No &
    Dim-16 WGRAD &
    Static &
    SR &
    None \\
    \midrule
    \textbf{Update 1} &
    1D &
    Yes &
    Dim-16 WGRAD &
    Static &
    SR &
    None \\
    \midrule
    \textbf{Update 2} &
    1D &
    Yes &
    Dim-128 WGRAD, DGRAD &
    Resampled &
    SR &
    None \\
    \midrule
    \textbf{Update 3} &
    1D &
    Yes &
    Dim-128 WGRAD, DGRAD &
    Resampled &
    MS-EDEN &
    W.T, X.T \\
    \midrule
    \textbf{\newrecipe} &
    1D &
    Yes &
    Dim-128 WGRAD, DGRAD &
    Resampled &
    MS-EDEN &
    W, X,  W.T, X.T  \\
    \bottomrule
  \end{tabularx}
\end{table*}

\paragraph{Questions.}
Our experiments are organized around the following questions:  

\begin{itemize}
    \item Do these new recipe techniques provide incremental benefits to model quality at scale? 
    \item Are selective high-precision layers necessary for long-horizon accuracy? 
    \item How do the benefits of unbiased quantization vary across backward-pass tensors? 
    \item Do trends with NVFP4 quantization reproduce across dense and MoE architectures?
\end{itemize}

\section{Experimental Results}

\subsection{Ablation Studies on Recipe Modifications}
We ablate each recipe modification on Nemotron 3 Nano, starting from the  \citet{nvidia2025pretraining} recipe and stacking techniques sequentially (Table \ref{recipe_table}). We refer to the final stacked recipe as \newrecipe. The first update replaces two-dimensional weight quantization with one-dimensional quantization, re-quantizing in the backward pass. The second adds a dimension-128 tiled Random Hadamard Transform to DGRAD and updates the WGRAD transform to use a consistent tile shape and a resampled sign vector throughout training. The third applies MS-EDEN to gradient tensors in place of stochastic rounding and adds Four Over Six scale selection to the remaining backward-pass tensors. The fourth adds Four Over Six to the forward-pass tensors. Each modification improves training loss (Figure \ref{fig:recipe_stacking}), yielding a 0.36 pp total improvement once fully stacked. The largest single gain comes from replacing stochastic rounding with MS-EDEN and adding Four Over Six on the backward tensors, which we attribute largely to MS-EDEN's ability to achieve unbiasedness at lower quantization-error cost than stochastic rounding \citep{panferov2026quartetii}. When running additional ablations with the MS-EDEN technique, we found that resampling the stochastic-rounding pattern for corrected scale factors, rather than reusing a fixed pattern, and independently randomizing the DGRAD RHT sign vector for each data-parallel replica, rather than once per global batch, yielded an additional 0.028 pp improvement in loss gap on the 30B-A3B model with a different optimizer configuration. The results in this work were obtained without these extra sources of stochasticity, suggesting improvements reported here are likely conservative, and we leave studying impacts of extra stochasticity as a subject of future work.

\begin{figure}[h!]
  \centering
  \includegraphics[width=\columnwidth]{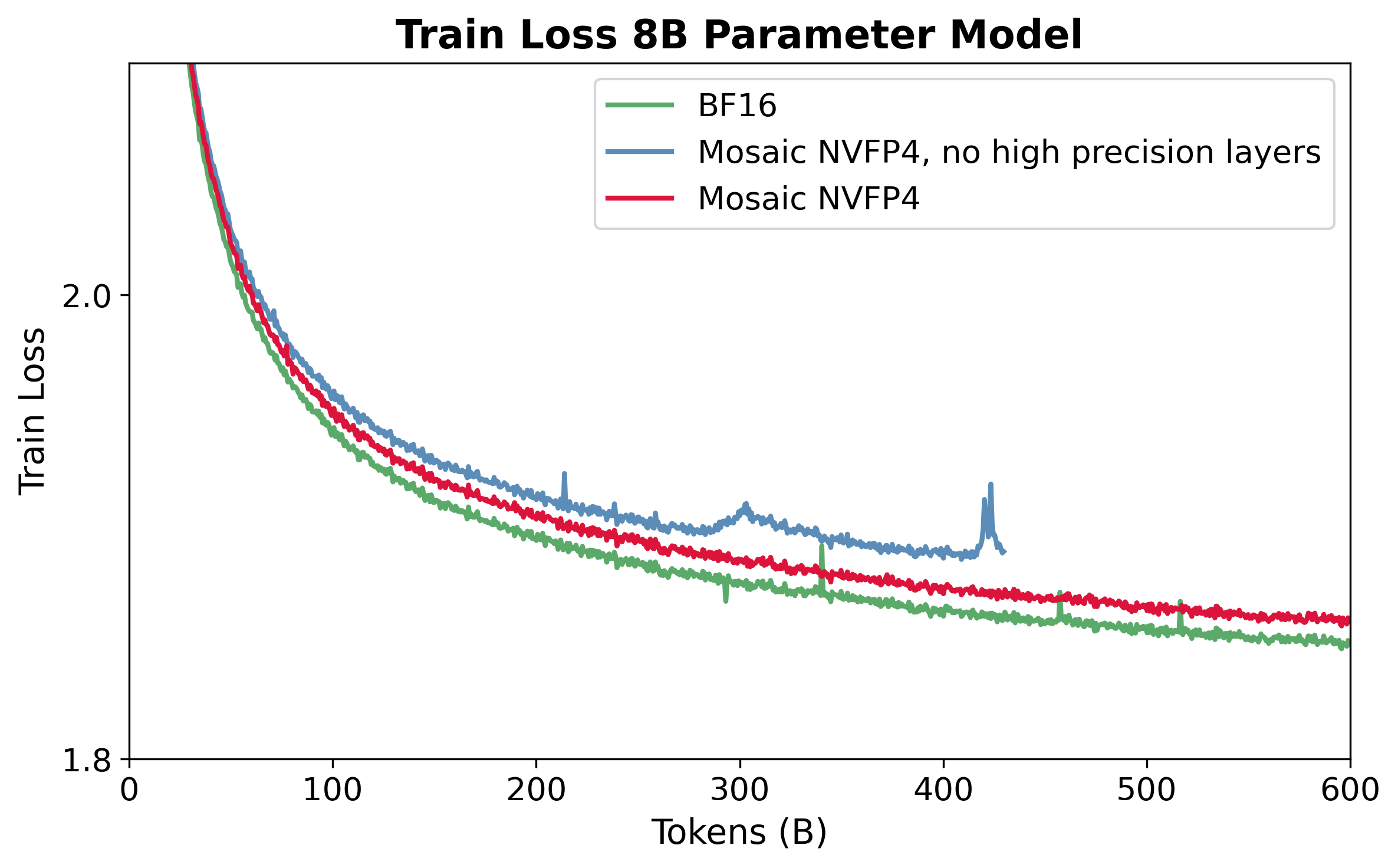}
  \caption{Stability study of the \newrecipe recipe with the last 15\% of the network kept in BF16 vs. all layers quantized on the Nemotron-H 8B model.}
  \label{fig:last_15_pct_bf16}
\end{figure}

\begin{figure*}[h!]
\centering
\includegraphics[
  width=\textwidth,
  trim=1pt 1pt 1pt 1pt,
  clip
]{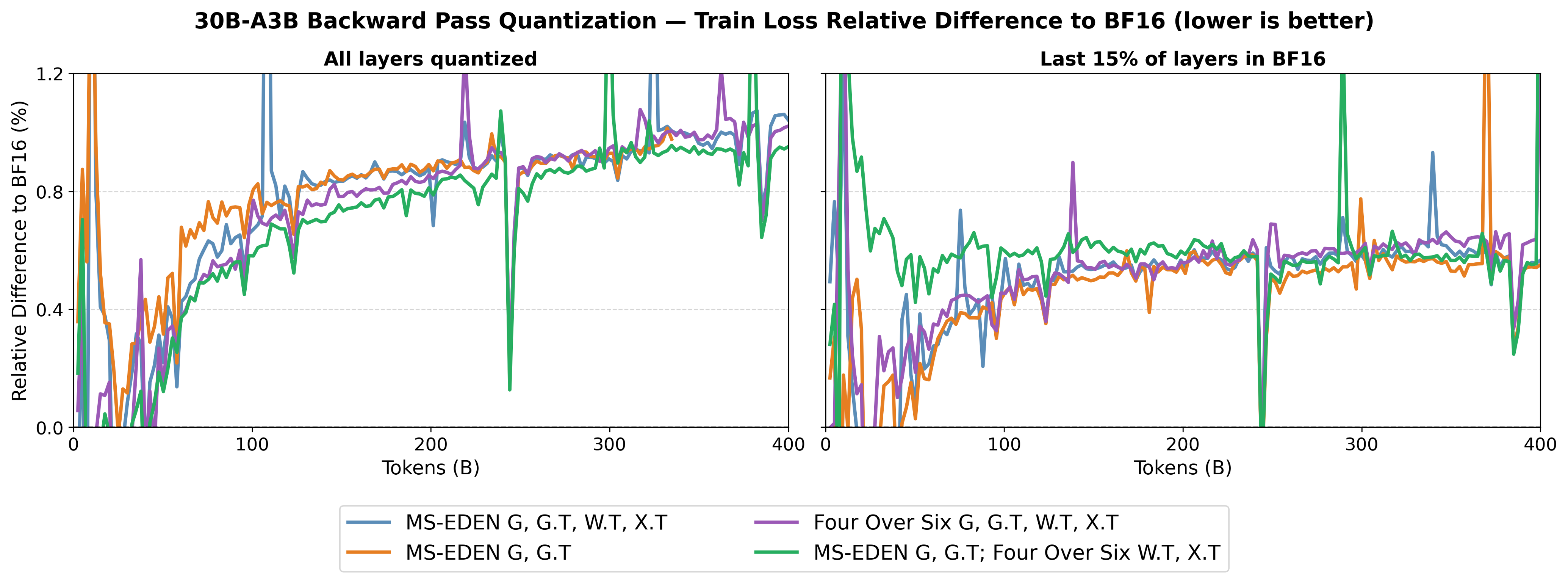}
  \caption{Different backward-quantization schemes on Nemotron 3 Nano, with all layers quantized (left) or the last 15\% of layers kept in BF16 (right). With all layers quantized, the configuration applying MS-EDEN to gradient tensors and Four Over Six scale selection to the remaining backward tensors, X.T and W.T, yields the lowest observed loss gap. At this model size and token horizon, the backward-pass techniques show no clear separation when the last 15\% of layers are kept in BF16.}
  \label{fig:backward_unbiased}
\end{figure*}

\subsection{Selective High-Precision Layers are Required}
Figure \ref{fig:last_15_pct_bf16} shows that the \newrecipe recipe diverges on the Nemotron-H 8B model past 300B tokens when all layers are quantized to NVFP4.

\begin{figure*}[h]
\centering
  \includegraphics[width=\textwidth]{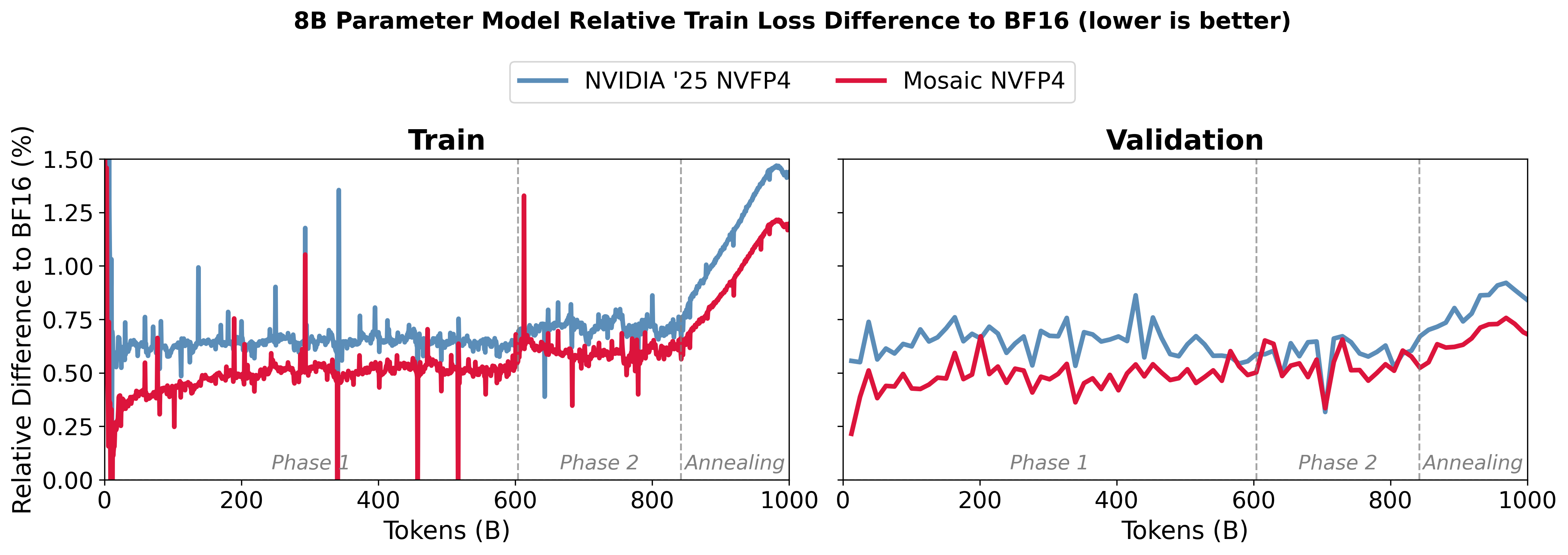}
  \caption{Percent relative train (left) and validation (right) loss difference compared to BF16 (lower is better) evaluated on the Nemotron-H 8B model to 1T tokens. The \newrecipe recipe has improved training and validation loss over most of the token horizon.}
  \label{fig:final_recipe_8B}
\end{figure*}

\subsection{Benefits of Backward-Pass Unbiasedness}
We studied backward pass quantization strategies on Nemotron 3 Nano. Recipes in this set of ablations informed Update 3 of Table \ref{recipe_table}. Each uses 1D weight quantization with re-quantization in the backward pass and dimension-128 tiled Random Hadamard Transforms on both WGRAD and DGRAD. We vary the use of MS-EDEN and Four Over Six on backward tensors, applying: i) MS-EDEN on all backward tensors, ii) MS-EDEN on gradients only, iii) Four Over Six on all backward tensors, and iv) MS-EDEN on gradients and Four Over Six on the remaining backward tensors (X.T, W.T). These recipes are tested with and without the last 15\% of the network kept in BF16. As shown in Figure \ref{fig:backward_unbiased}, applying MS-EDEN to all tensors in backward GEMMs provides no clear benefit over applying MS-EDEN only to gradients. When all layers are quantized, the recipe applying MS-EDEN to G and G.T and Four Over Six to X.T and W.T yields the lowest observed loss gap among the tested recipes, although the separation is small. This may indicate that bias correction is most important for the gradient tensors, supporting results from \citep{nvidia2025pretraining}.

\subsection{The Best Stacked Recipe}
Both the \newrecipe and \nvrecipe recipes were used to train Nemotron 3 Nano (30B-A3B) (Figure \ref{fig:recipe_stacking}) and Nemotron-H 8B (Figure \ref{fig:final_recipe_8B}) to 1T tokens, keeping the last 15\% of the network in BF16 for the entirety of training. On the dense model, the version trained with the \newrecipe{} recipe finishes training with less than a 1.2\% relative train loss gap from BF16 and improves upon the previous loss gap by 0.24 percentage points. The benefit of \newrecipe is most pronounced in the beginning of training and then during the annealing stage. Train and validation loss gap dynamics differ slightly due to dataset differences, but the gap between recipes is consistent across both. The switch from Phase 1 to Phase 2 data introduces a transiently higher loss gap for the \newrecipe{} recipe, though sustained improvements are  visible again after training has progressed in Phase 2. Both recipes show an increased loss gap from BF16 during the learning rate annealing phase. On the MoE model, the \newrecipe{} recipe shows sustained improvement throughout the token horizon, starting after the first 75B tokens (Figure \ref{fig:recipe_stacking}). The model trained with the \newrecipe recipe finishes training with less than a 0.81\% relative train loss gap from BF16 and improves upon the previous loss gap by 0.36 percentage points. The training curriculum is restricted to a single phase of data so the only change in loss gap trajectory occurs during the learning rate annealing, which impacts both recipes equally (Figure \ref{fig:recipe_stacking}). These differences exceed the run-to-run variability observed across three repeated runs at the 1T-token horizon (Appendix~\ref{app:loss-variability}).

\section{Discussion}
 Prior work has shown that a close variant of the \newrecipe recipe can converge on small models over short token horizons without keeping any layers in high precision \citep{panferov2026quartetii}. However, at scale these NVFP4 recipes still require high-precision layers, suggesting a representation failure that the error-reduction techniques studied in this work cannot address. This is shown most directly on the 8B parameter model trained to 300B tokens in Figure \ref{fig:last_15_pct_bf16}, and further supported by Figure \ref{fig:backward_unbiased}, where all recipes, regardless of the backward-tensor technique used, show a loss gap that grows throughout training.

 Figure \ref{fig:recipe_stacking} shows that the relative ranking of recipes does not stabilize until well past 150B tokens of training. Notably, weaker recipes often initially appear better than the strongest ones. These early changes in ordering can be attributed to differences in the randomly sampled sign vectors used by the Random Hadamard Transforms, noise in training, or convergence properties of individual recipes. Only after enough tokens have been seen do the rankings settle. This view suggests that short NVFP4 pretraining ablations are useful for detecting catastrophic instability, but they are unfortunately not sufficient for deciding whether a new training algorithm will preserve accuracy at scale. For FP4 pretraining methods, where the intended benefit is to make long training runs cheaper, this late-emerging failure mode is especially important.

 The separation between techniques applied to backward tensors is more visible in a less stable training regime. Figure \ref{fig:backward_unbiased} shows a slight separation when all layers are quantized, but indistinguishable curves once the last 15\% of the network is kept in BF16. For our full training runs, we adopted a recipe that addresses bias in the gradients and reduces quantization error in the subsequent backward tensors, applying targeted interventions to each tensor type. A pragmatic alternative entails applying MS-EDEN to gradients without additional techniques on X.T and W.T, likely without significant cost to model quality in the stable regime, as Figure \ref{fig:backward_unbiased} suggests.

The \newrecipe recipe incrementally improves upon the \nvrecipe recipe, remains robust across the two model sizes, datasets, and architectures explored in this work, and partially closes the training-loss gap to BF16 at both scales.

\section{Limitations}

This paper focuses on medium-scale dense and MoE language models and on NVFP4 training. We do not yet establish whether the same conclusions hold for even larger production runs, other modalities, post-training, or other FP4 formats. We leave this as future work.

\bibliography{bib_final}
\bibliographystyle{icml2026}

\appendix
\onecolumn

\section{Appendix}

\subsection{Model and Training Parameters}
\label{model_and_training_params}
Model and training parameters are summarized in Tables \ref{model_params} and \ref{train_params}. In our experiments, both models were trained using AdamW with the Warmup-Stable-Decay learning-rate schedule, with annealing over the final 20\% of the token horizon for Nemotron 3 Nano and the final 15\% for Nemotron-H 8B. Each model was trained to a maximum of 1T tokens; some ablations were stopped early in the schedule. Weight decay of 0.1 was used for all models. Attention layers used Grouped Query Attention. All models have a sequence length of 8192.

\begin{table}[h]
  \centering
  \caption{Model parameters for each experimental setup. Active parameter counts exclude embeddings. $^\dagger$ Nemotron 3 Nano uses MoE layers.}
  \begin{tabular}{|c|c|c|c|c|c|}
    \hline
    Model & No. Mamba-2 Layers & No. Attention Layers & \shortstack{No. FFN/MoE$^\dagger$\\Layers} & No. KV Heads & \shortstack{Active/ Total\\Parameters}\\
    \hline
    Nemotron 3 Nano & 23 & 6 & 23 & 2 & 3.2B / 31.6B\\
    \hline
    Nemotron-H & 24 & 4 & 24 & 8 & 8B \\
    \hline
  \end{tabular}
  \label{model_params}
\end{table}

\begin{table}[h]
  \centering
  \caption{Training parameters for each experimental setup.}
  \begin{tabular}{|c|c|c|c|}
    \hline
    Model & Batch Size & Min LR & LR\\
    \hline
    Nemotron 3 Nano & 3072 & 1e-5 & 1e-3\\
    \hline
    Nemotron-H & 768 & 8e-6 & 8e-4 \\
    \hline
  \end{tabular}
  \label{train_params}
\end{table}

\subsection{Downstream Task Accuracy and Run-to-Run Variability}

To determine whether differences in training and validation loss gaps and downstream task accuracy exceed expected
run-to-run variability, we trained the \nvrecipe recipe three times and evaluated each
resulting model on the same downstream tasks. 

\subsubsection{Loss Gap Variability}
\label{app:loss-variability}
We measure run-to-run loss variability as
$\max_{a \neq b} |L_a-L_b|/L_b$ across the three repeated runs, where $L$ is the loss measured at the 1T-token horizon.
For the Nemotron 3 Nano model, this variability is approximately \(0.06\%\) in training loss and \(0.04\%\) in validation loss; for the Nemotron-H 8B model, it is \(0.027\%\) and \(0.064\%\), respectively. Loss differences substantially exceeding these ranges are unlikely to be explained by run-to-run variability alone.

\subsubsection{Downstream Task Accuracy}
For model $m$ and task $t$, we
estimate the relative noise tolerance as
\begin{equation}
    \epsilon_{m,t} =
    \frac{\max_i s_{m,t,i} - \min_i s_{m,t,i}}
         {\max_i s_{m,t,i}},
\end{equation}
where $s_{m,t,i}$ is the score from repeated run $i$. We then calculate the
relative difference between \newrecipe and the \nvrecipe baseline:
\begin{equation}
    \Delta_{m,t} =
    \frac{s_{m,t}^{\mathrm{Mosaic}} -
          s_{m,t}^{\mathrm{NV25}}}
         {s_{m,t}^{\mathrm{NV25}}}.
\end{equation}
A positive $\Delta_{m,t}$ indicates that \newrecipe improves accuracy.
Differences with $\lvert\Delta_{m,t}\rvert > \epsilon_{m,t}$
are considered outside the corresponding noise reference. Because this estimate is based on only three runs, the measured range may understate the true run-to-run variability.

For each aggregate category $C$, we report the mean of
its constituent task scores and compute the relative difference
between the resulting NVIDIA~'25 NVFP4 and Mosaic NVFP4 category scores.
This aggregate relative difference can equivalently be written as a
weighted average of the task-level relative differences:
\[
\Delta_{m,C}
=
\sum_{t \in C} w_{m,t}\Delta_{m,t}
=
\frac{
\bar{s}^{\,\mathrm{Mosaic}}_{m,C}
-
\bar{s}^{\,\mathrm{NV25}}_{m,C}
}{
\bar{s}^{\,\mathrm{NV25}}_{m,C}
},
\qquad
w_{m,t}
=
\frac{s^{\mathrm{NV25}}_{m,t}}
{\sum_{j \in C}s^{\mathrm{NV25}}_{m,j}}.
\]
Here, $w_{m,t}$ is the normalized contribution of task $t$ to the
aggregate NVIDIA~'25 NVFP4 score. We use the same weights to combine
the task-level noise tolerances:
\[
\widetilde{\epsilon}_{m,C}
=
\sum_{t \in C}w_{m,t}\epsilon_{m,t}.
\]

We use $\widetilde{\epsilon}_{m,C}$ as a score-aligned
aggregate noise reference.

\begin{table*}[h]
\centering
\caption{
Downstream-task scores and measured run-to-run variability.
For individual tasks, the noise reference is the relative range across
three repeated runs. For aggregate rows, it is an average weighted by the \nvrecipe task scores. Boldface denotes differences outside the corresponding
noise reference.
}
\label{tab:downstream_noise}
\resizebox{\textwidth}{!}{%
\begin{tabular}{lrrrr}
\toprule
\textbf{Task}
& \textbf{\shortstack{NVIDIA '25\\NVFP4 (0--100)}}
& \textbf{\shortstack{Mosaic\\NVFP4 (0--100)}}
& \shortstack{\textbf{Noise}\\\textbf{Reference (\%)}}
& \textbf{$\boldsymbol{\Delta}$ \textbf{(\%)}} \\
\midrule

\multicolumn{5}{l}{\textbf{Nemotron 3 Nano (30B-A3B)}} \\
\midrule

\rowcolor{gray!12}
\multicolumn{5}{l}{\textbf{General Knowledge}} \\
MMLU, 5-shot
& 67.5 & 68.4 & 0.9 & \textbf{+1.3} \\
MMLU-Pro, 5-shot CoT
& 45.0 & 47.1 & 6.3 & +4.7 \\

\rowcolor{gray!12}
\textbf{Math Average}
& 79.4 & 79.2 & 4.6 & -0.3 \\
\multicolumn{5}{l}{\textit{(Math 500, 4-shot; GSM8K, 8-shot CoT)}} \\

\rowcolor{gray!12}
\textbf{Common Sense Average}
& 77.6 & 77.1 & 1.4 & -0.6 \\
\multicolumn{5}{l}{\textit{(ARC Challenge, 25-shot; WinoGrande, 0-shot;
HellaSwag, 0-shot)}} \\

\rowcolor{gray!12}
\textbf{Code Average}
& 57.1 & 56.6 & 6.8 & -0.9 \\
\multicolumn{5}{l}{\textit{(HumanEval, average Pass@1;
MBPP Sanitized, 3-shot)}} \\

\rowcolor{gray!12}
\multicolumn{5}{l}{\textbf{Reading Comprehension}} \\
RACE, 0-shot
& 80.8 & 81.1 & 0.7 & +0.4 \\

\midrule
\multicolumn{5}{l}{\textbf{Nemotron-H 8B}} \\
\midrule

\rowcolor{gray!12}
\multicolumn{5}{l}{\textbf{General Knowledge}} \\
MMLU, 5-shot
& 61.0 & 60.3 & 2.2 & -1.1 \\
MMLU-Pro, 5-shot CoT
& 29.7 & 27.6 & 9.8 & -7.1 \\

\rowcolor{gray!12}
\textbf{Math Average}
& 51.3 & 52.3 & 4.6 & +1.9 \\
\multicolumn{5}{l}{\textit{(Math 500, 4-shot; GSM8K, 8-shot CoT)}} \\

\rowcolor{gray!12}
\textbf{Common Sense Average}
& 76.5 & 75.9 & 1.1 & -0.8 \\
\multicolumn{5}{l}{\textit{(ARC Challenge, 25-shot; WinoGrande, 0-shot;
HellaSwag, 0-shot)}} \\

\rowcolor{gray!12}
\textbf{Code Average}
& 43.4 & 43.0 & 5.6 & -0.9 \\
\multicolumn{5}{l}{\textit{(HumanEval, average Pass@1;
MBPP Sanitized, 3-shot)}} \\

\rowcolor{gray!12}
\multicolumn{5}{l}{\textbf{Reading Comprehension}} \\
RACE, 0-shot
& 77.7 & 78.8 & 1.1 & \textbf{+1.4} \\

\bottomrule
\end{tabular}%
}
\end{table*}
\end{document}